\documentclass[journal,twocolumn]{IEEEtran}

\usepackage{amsmath}
\usepackage{algorithm}
\usepackage{algpseudocode}
\usepackage{setspace}
\usepackage{color,hyperref}
\usepackage{bbding} %首先在导言区调用bbding包

\usepackage{graphicx}
\usepackage{amsmath,amssymb} % define this before the line numbering.
\usepackage{epsfig}
\usepackage{amsmath}
\usepackage{amssymb}
\usepackage{colortbl}
\usepackage{soul}
\usepackage{cite}
\usepackage{multirow}
\usepackage{pifont}
\usepackage{bm}

\usepackage{graphicx}
\usepackage{float}

\usepackage{caption}
\usepackage{subcaption}

\newcommand{\xmark}{\ding{55}}
\newcommand{\rmark}{\ding{52}}

\definecolor{mygray}{gray}{.9}

\newcolumntype{I}{!{\vrule width 1.2pt}}

\newlength\savedwidth
\newcommand\whline{\noalign{\global\savedwidth\arrayrulewidth
		\global\arrayrulewidth 1.25pt}%
	\hline
	\noalign{\global\arrayrulewidth\savedwidth}}

\ifCLASSINFOpdf
\else
\fi

\definecolor{darkblue}{rgb}{0.0,0.0,1.0}
\hypersetup{colorlinks,breaklinks,
	linkcolor=red,urlcolor=magenta,
	anchorcolor=blue,citecolor=green}

\begin{document}
\setul{}{1.5pt}

\title{MAFNet: A Multi-Attention Fusion Network \\ for RGB-T Crowd Counting}

\author{
        Pengyu~Chen,~\IEEEmembership{Student Member,~IEEE,} 
        Junyu Gao,~\IEEEmembership{Member,~IEEE,} 
        Yuan Yuan,~\IEEEmembership{Senior Member,~IEEE,}
        and \\
        Qi Wang,~\IEEEmembership{Senior Member,~IEEE}
    
    \thanks{
    
    Pengyu Chen is with the School of Computer Science, and the School of Artificial Intelligence, Optics and Electronics (iOPEN), Northwestern Polytechnical University, Xi’an 710072, P. R. China. (E-mail: chenpengyu2000222@gmail.com).
    
    Junyu Gao, Yuan Yuan and Qi Wang are with the School of Artificial Intelligence, Optics and Electronics (iOPEN), Northwestern Polytechnical University, Xi’an 710072, P. R. China. (E-mail: gjy3035@gmail.com; y.yuan1.ieee@gmail.com; crabwq@gmail.com).
    }
}

\markboth{{IEEE} Transactions on XXX}%
{Shell \MakeLowercase{\textit{et al.}}: Bare Demo of IEEEtran.cls for Journals}
%make the title area
\maketitle

\begin{abstract}
% RGB-thermal (RGB-T) crowd counting is a challenging task, which takes advantage of the insensitivity of thermal images to illumination variations to provide complementary information for RGB images to improve the reliability of the RGB-based methods in scenes with low-illumination or similar backgrounds. 
RGB-Thermal (RGB-T) crowd counting is a challenging task, which uses thermal images as complementary information to RGB images to deal with the decreased performance of unimodal RGB-based methods in scenes with low-illumination or similar backgrounds. Most existing methods propose well-designed structures for cross-modal fusion in RGB-T crowd counting. However, these methods have difficulty in encoding cross-modal contextual semantic information in RGB-T image pairs. Considering the aforementioned problem, we propose a two-stream RGB-T crowd counting network called Multi-Attention Fusion Network (MAFNet), which aims to fully capture long-range contextual information from the RGB and thermal modalities based on the attention mechanism. Specifically, in the encoder part, a Multi-Attention Fusion (MAF) module is embedded into different stages of the two modality-specific branches for cross-modal fusion at the global level. In addition, a Multi-modal Multi-scale Aggregation (MMA) regression head is introduced to make full use of the multi-scale and contextual information across modalities to generate high-quality crowd density maps. Extensive experiments on two popular datasets show that the proposed MAFNet is effective for RGB-T crowd counting and achieves the state-of-the-art performance. 

% RGB-T Crowd Counting incorporates information from RGB and thermal images to more accurately estimate the number of people in challenging situations like dark environments and complex backgrounds. However, the existing RGB-T crowd counting methods use the multi-modal fusion module produced by convolutional neural networks (CNNs), which is incapable of establishing long-range contexts dependencies in pixel-rich image data. Considering the aforementioned problem, a two-stream encoder-decoder network MAFNet is proposed for RGB-T crowd counting tasks. Specifically, in the encoder part, the well-designed Multi-Attention Fusion (MAF) module is embedded into some different stages of the two modality-specific branches for cross-modal information interaction and fusion at a global level. In the decoder part, the Pyramid Feature Aggregation (PFA) module is introduced to integrate multi-scale features from multiple modalities and the Multi-Scale Dilated Convolution (MDC) module to strengthen the integrated features to generate accurate regression density maps. Experiments results on two popular datasets show that the MAFNet is effective for RGB-T crowd counting tasks and achieves the state-of-the-art performance. 

% The source code will be publicly available at \textit{\url{https://github.com/gaiyi7788/MAFNet}}.

\end{abstract}


\section{INTRODUCTION}
\begin{figure}[!t]
\centering
\includegraphics[width=0.45\textwidth]{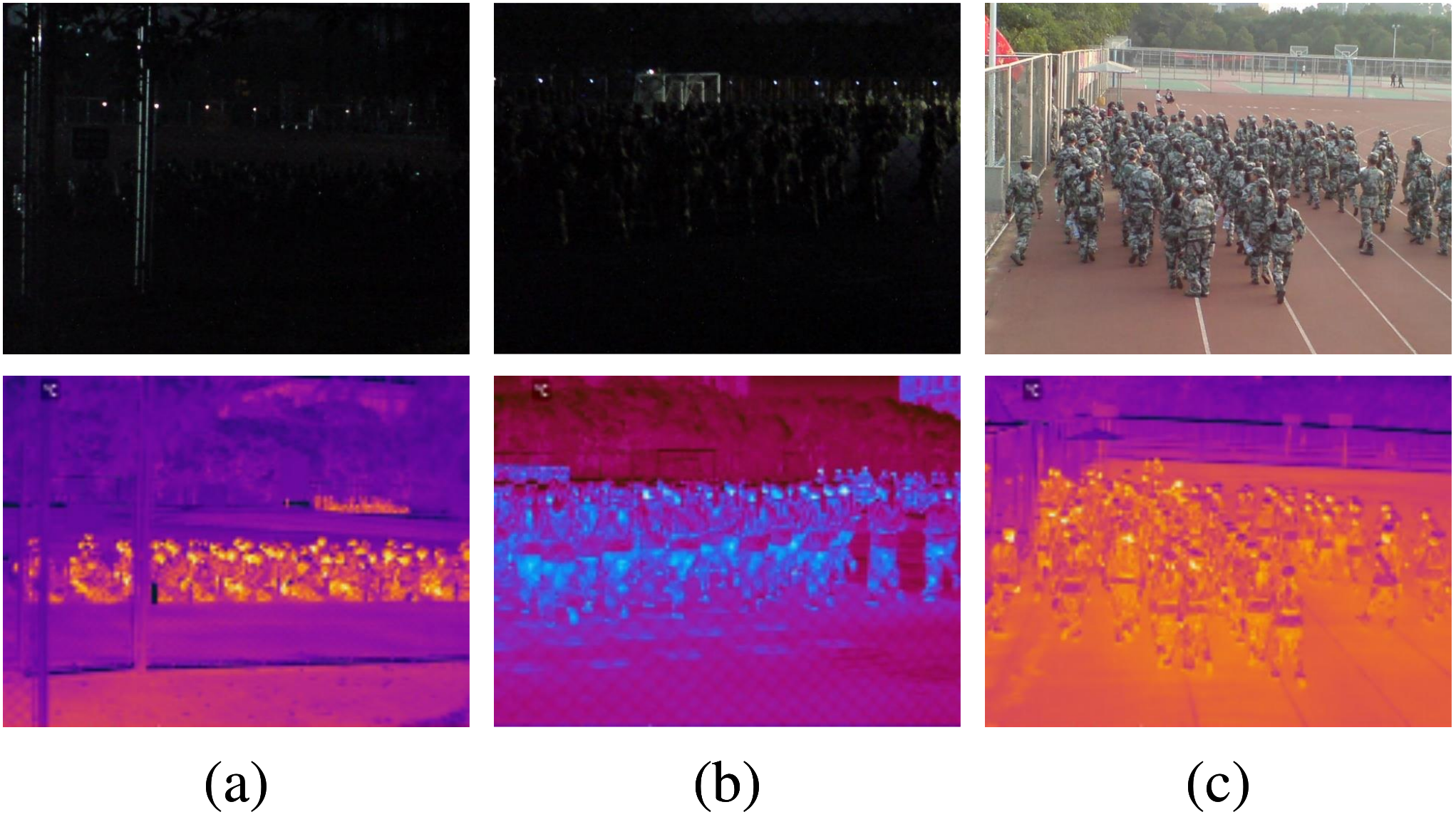}
\caption{(a) and (b): two pairs of RGB and thermal images in dark illumination. (c): a pair of RGB and thermal images in bright illumination.}
\label{rgbt_example}
\end{figure}

% \IEEEPARstart{A}{s} a research hot topic of computer vision, crowd counting  \cite{ccsurvy2020} \cite{BeyondCounting} aims to identify the number of individuals in images and plays an important role in intelligent transportation, city management and security surveillance. 
\IEEEPARstart{I}{n} recent years, with the increase in population, crowd counting has become an important topic in the field of intelligent surveillance \cite{zhang2017understanding, xiong2017spatiotemporal, dosovitskiy2020image}, human swarm analysis \cite{rosenberg2015human, 7780278, yu2020intelligent} and  public security \cite{helbing2002simulation, sang2019improved}, where the target is to accurately estimate the number of people in images. Current crowd counting methods \cite{lempitsky2010learning, zhang2016single, li2018csrnet,  jiang2019learning, gao2019scar, gao2019pcc, wang2019learning, gao2020feature, wang2021neuron, gao2021domain} mainly rely on visual features of RGB images. However, RGB images are susceptible to illumination variations, which limits the application of RGB-based methods in scenes with low-illumination or similar backgrounds. Different from RGB images, thermal images are insensitive to illumination variations and have a powerful penetration to fog and smog. Recently, some researches introduce thermal images as complementary information to RGB images in computer vision tasks. As shown in Fig. \ref{rgbt_example}-(a, b), crowds in RGB images are invisible in dark conditions, whereas the outlines of crowds in the thermal images are clearly visible. Meanwhile, thermal images have low resolution and suffer from thermal crossover problems to confuse different objects. In contrast, RGB images have high resolution and rich semantic information, which could cover the above shortcomings of thermal images. As shown in Fig. \ref{rgbt_example}-(c), crowds in thermal images are difficult to distinguish because of the thermal crossover, while the crowds are clearly visible in RGB images in bright illumination. Thus, it is necessary to study the RGB-Thermal cross-modal fusion problem in crowd counting.

% fuse the information of both RGB and thermal modalities for crowd counting. 

% Due to the complementary properties of the modalities, full information interaction between two modalities could improve the performance of the crowd counting methods.

Recently, RGB-Thermal (RGB-T) crowd counting has received much attention, and several datasets and methods are proposed. The existing methods \cite{peng2020rgb, liu2021cross, zhang2021mmccn, tang2022tafnet} manually propose well-designed structures for RGB-T fusion based on convolutional neural networks (CNNs) and show exceptional performance on DroneRGBT \cite{peng2020rgb} and RGBT-CC \cite{liu2021cross} datasets. Peng \textit{et al.} \cite{peng2020rgb} first propose an end-to-end framework named Multi-Modal Crowd Counting Network (MMCCN) with an adaptive fusion module. Based on this, Zhang \textit{et al.} \cite{zhang2021mmccn} propose I-MMCCN, which improves MMCCN by introducing a hard example mining module and a novel block-averaged absolute error loss. Besides, Liu \textit{et al.} \cite{liu2021cross} propose an Information Aggregation-Distribution Module (IADM) for cross-modal representation learning. Further, Tang \textit{et al.} \cite{tang2022tafnet} propose a three-stream adaptive fusion network (TAFNet) to extract the combination features of RGB-T through an additional stream. However, purely CNN-based methods have difficulty in adequately capturing long-range contextual information \cite{cordonnier2019relationship, zhao2021RSGN} from both modalities on account of the small perception field of convolution operations, which limits the effectiveness of cross-modal fusion.

% Additional, the inductive bias of the convolution operations reduce the generalization ability of the models that constrains the performance of the CNN-based methods to a ceiling.

% Recently, owing to the powerful cross-modal global modeling capability, attention-based \cite{vaswani2017attention} methods \cite{tan2019lxmert}, \cite{kim2021vilt}, \cite{radford2021learning}, \cite{dou2021empirical}, \cite{botach2021end} have developed rapidly in the field of multi-modal fusion. These models convert the input data from different modalities into sequences uniformly and apply an attention mechanism to obtain the long-range correlation and global information of the original images. A natural idea is to introduce the attention mechanism into the RGB-T crowd counting model to fully tap the multi-modal information and alleviate the problem of the small global receptive field of the CNN-based methods.

In recent works, the attention mechanism \cite{vaswani2017attention} has demonstrated the usefulness in cross-modal fusion to obtain long-range correlation and global information of the two modalities \cite{tan2019lxmert,kim2021vilt,radford2021learning}. To tackle the limitation of the CNN-based RGB-T crowd counting methods, we propose a two-stream framework called Multi-Attention Fusion Network (MAFNet). Specifically, in the encoder part, two VGG19 \cite{simonyan2014very} backbones are used for modality-specific representation learning, and the attention-based Multi-Attention Fusion (MAF) modules are embedded in the backbones to capture long-range contextual information for cross-modal fusion in a hierarchical manner. Besides, considering that the feature maps may lose crucial information in the process of down-sampling, we introduce a Multi-modal Multi-scale Aggregation (MMA) regression head. The extracted features with different scales from different modalities are fed to the MMA regression head to make full use of the multi-scale and contextual information across modalities to generate high-quality crowd density maps.

% owing to the powerful cross-modal global modeling capability of attention mechanism \cite{cordonnier2019relationship}, \cite{prakash2021multi}.

%\vspace*{0.5\baselineskip} 
Our main contributions are summarized as follows:
%\vspace*{0.5\baselineskip} 

% A two-stream encoder-decoder structure network called Multi-Attention Fusion Network (MAFNet) is proposed to utilize complementary information from RGB and thermal modalities for crowd counting.

\begin{itemize}
  \item [1)] 
A two-stream RGB-T crowd counting framework termed Multi-Attention Fusion Network (MAFNet) is proposed with an attention-based Multi-Attention Fusion (MAF) module to fully capture long-range contextual information from both modalities in the encoder part.

\vspace*{0.5\baselineskip} 
  \item [2)]
A Multi-modal Multi-scale Aggregation (MMA) regression head is introduced to make full use of the multi-scale and contextual information across modalities to generate high quality crowd density maps.

\vspace*{0.5\baselineskip} 
  \item [3)]
Experiments show that the proposed MAFNet achieves the state-of-the-art performance on two RGB-T crowd counting datasets, RGBT-CC and DroneRGBT.
\end{itemize}

\section{RELATED WORK}
The discussion of related works has been divided into two categories: (A) RGB-based crowd counting and (B) multi-modal crowd counting.

\subsection{RGB-Based Crowd Counting}
The development of crowd counting has gone through three stages: detection-based, regression-based, and density estimation-based methods.

\textbf{Detection-based methods.} The early crowd counting methods are mainly based on object detection \cite{topkaya2014counting, viola2005detecting, li2008estimating, leibe2005pedestrian, felzenszwalb2009object, wang2022crowd}. After the detection of individuals, the number of targets is counted to obtain the total number of people. Such algorithms perform well in scenes with sparse crowds and no occlusion, but poorly in dense crowds.

\textbf{Regression-based methods.} Due to the limitations of detection-based methods, researchers propose regression-based crowd counting methods \cite{chan2008privacy, chan2009bayesian, idrees2013multi}. These methods first extract global features or local features of the input image, and then use regression methods to learn a mapping function to obtain the crowd counts. However, the extracted low-level features may ignore the critical features of the targets and such methods cannot obtain the spatial distribution of the crowds, limiting the application in real scenes.

\textbf{Density estimation-based methods.} The methods based on density estimation~\cite{lempitsky2010learning, zhang2016single, li2018csrnet, sindagi2017generating, gao2019scar, liu2019context, gao2019pcc, yan2019perspective, yang2020reverse, wang2019learning, gao2020feature, gao2021domain, wei2021scene, tian2021cctrans, lin2022boosting} use spatial information to estimate the crowd density maps, which are the current mainstream crowd counting methods. The crowd density map not only reflects the spatial distribution of the crowds in the image, but also its integral value is the crowd counts. Zhang \textit{et al.} \cite{zhang2016single} propose a Multi-column Convolutional Neural Network (MCNN) for crowd counting, which utilizes filters with different sizes of receptive fields to extract features. Li \textit{et al.} \cite{li2018csrnet} propose a Congested Scene Recognition Network (CSRNet) that uses dilated convolution in the backend to improve the receptive field and enhances feature extraction capability. Recently, some works improve the crowd counting performance by introducing multi-scale information \cite{sindagi2017generating, gao2019scar, liu2019context} and perspective estimation \cite{gao2019pcc, yan2019perspective, yang2020reverse}. To alleviate the problem of difficult collection and annotation of crowd counting datasets, some works \cite{wang2019learning, gao2020feature, gao2021domain} explore the domain-adaptive crowd counting from the synthetic datasets to the real-world. In addition, with the Vision Transformer (ViT) \cite{dosovitskiy2020image} first applying the transformer structure for vision tasks, many transformer-based \cite{vaswani2017attention} crowd counting methods \cite{wei2021scene, tian2021cctrans, lin2022boosting} have been proposed with outstanding performance. 

% With the development of vision-transformers, researchers are considering introducing the vision-transformer to crowd counting tasks for global context modeling to improve the performance of crowd counting algorithms. 

\begin{figure*}[!t]
\centering
\includegraphics[width=0.9\textwidth]{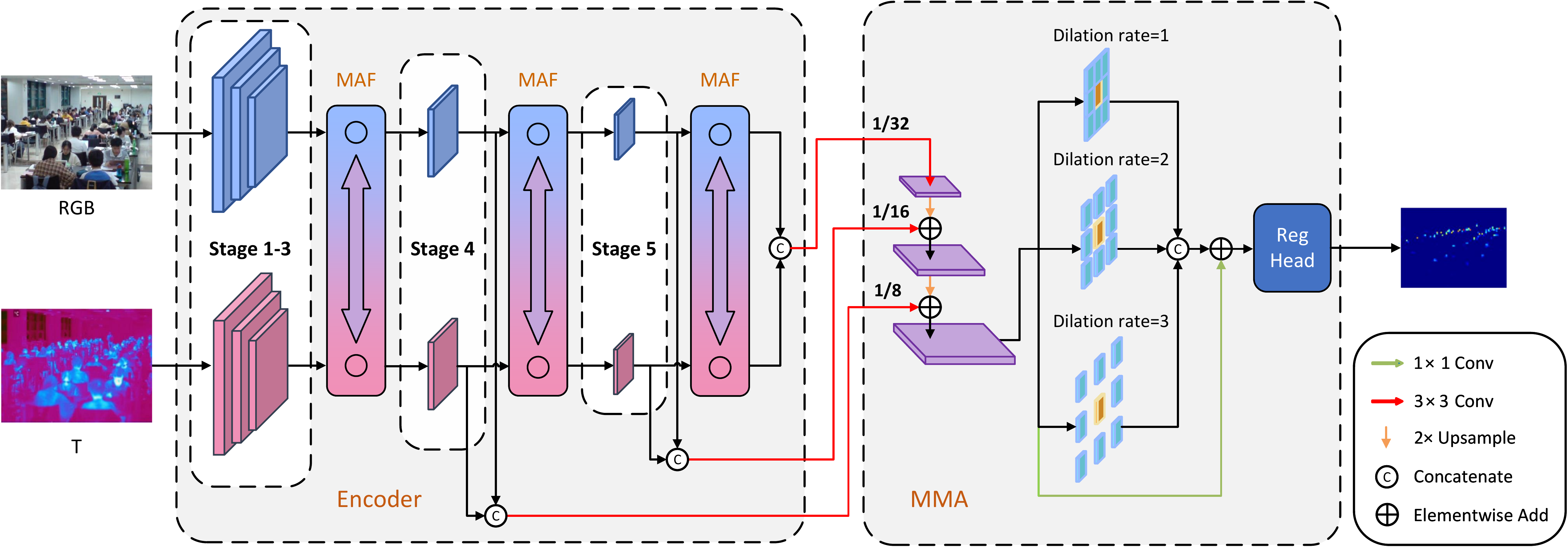}
\caption{The flowchart of the proposed two-stream network Multi-Attention Fusion Network (MAFNet) for RGB-T crowd counting. Specifically, the encoder of the network consists of two VGG19 backbones for modality-specific representation learning and three Multi-Attention Fusion (MAF) modules, which are embedded between the stages of the backbones. A Multi-modal Multi-scale Aggregation (MMA) regression head is used for generating predicted crowd density maps.}
\label{structure}
\end{figure*}

\subsection{Multi-Modal Crowd Counting}
Multi-modal crowd counting aims at improving the performance of crowd counting methods by comprehending multi-modal data through multi-modal representation learning \cite{guo2019deep, bayoudh2021survey, 9578480, Fu2020JLDCF, Fu2021siamese}. Multi-modal crowd counting methods mainly include three research directions: the methods based on RGB and depth images (RGB-D), based on auditory and visual information (Audio-Visual), and based on RGB and thermal images (RGB-T).

\textbf{RGB-D crowd counting methods.} The depth image reflects the scale information of the image, which facilitates the detection of small-scale targets for crowd counting. Bondi \textit{et al.} \cite{bondi2014real} propose a RGB-D crowd counting dataset MICC and use depth images to segment people and localize candidate heads. Arteta \textit{et al.} \cite{arteta2016counting} introduce depth information and enhance crowd density estimation performance by foreground and background segmentation. Song \textit{et al.} \cite{song2017depth} use the Kinect sensor to obtain depth images and propose a depth region suggestion network for counting. Lian \textit{et al.} \cite{lian2019density}, \cite{lian2021locating} propose a strategy to generate crowd depth-adaptive density maps, and a real-world RGB-D crowd counting dataset called ShanghaiTechRGBD and a synthetic dataset ShanghaiTechRGBD-syn are collected. Yang \textit{et al.} \cite{yang2019deccnet} propose a Bidirectional Cross-modal Attention (BCA) mechanism to focus on crowded regions in images though depth information. Li \textit{et al.} \cite{li2022rgb} fuse the RGB and depth images by cycle-attention and optimize the counting model from both fine pixel-aware and coarse pyramid region-aware perspectives. However, the depth information is easily disturbed and poorly robust to illumination changes and occlusions in practical applications. 

\textbf{Audio-Visual crowd counting methods.} Researchs in the field of neurobiology have shown that visual and auditory information are widely used as perceptual mediators of human beings. Several researches introduce auditory information as an auxiliary cue for crowd counting tasks. Hu \textit{et al.} \cite{hu2020ambient} propose an estimation model that jointly learns visual and audio modalities, and release a large-scale audiovisual crowd counting dataset DISCO. Sajid \textit{et al.} \cite{sajid2021audio} propose an audio-visual multi-task network based on the transformer structure to achieve better pattern association and efficient feature extraction. Hu \textit{et al.} \cite{hu2021avmsn} propose an Audio-Visual Multi-Scale Network (AVMSN) to model unconstrained visual and auditory sources for crowd counting. Nevertheless, auditory information is easily disturbed by the surrounding noise in a noisy scenario.

\textbf{RGB-T crowd counting methods.} Thermal images are insensitive to illumination variations and have strong ability to penetrate some particulate matters such as smog and fog, which could be used as complementary information to RGB images for crowd counting. Peng \textit{et al.} \cite{peng2020rgb} propose the DroneRGBT dataset, which is the first RGB-T crowd counting dataset based on drone view, and propose the MMCCN network to learn both modal information from RGB and thermal images simultaneously. Based on this, Zhang \textit{et al.} \cite{zhang2021mmccn} propose an improved model of MMCCN, the I-MMCCN network, which introduces a hard example mining module and a novel block average absolute error loss to the MMCCN network for performance enhancement. Besides, Liu \textit{et al.} \cite{liu2021cross} propose a large-scale RGB-T crowd counting dataset RGBT-CC and a two-stream cross-modal representation learning framework. Tang \textit{et al.} \cite{tang2022tafnet} propose a three-stream adaptive fusion network named TAFNet, which extracts the combination features of RGB and thermal images through an additional stream. After introducing the thermal information, the performance of crowd counting methods in low-illumination and similar backgrounds is effectively improved. However, the existing RGB-T crowd counting methods use CNN-based cross-modal fusion structure, which makes it difficult to obtain long-range contextual information across modalities due to the small perception field. Unlike the previous methods, we adopt a multi-modal fusion method based on the attention mechanism, which can effectively focus on the global and contextual information between the modalities. 

\section{PROPOSED METHOD}

In this section, we introduce the full structure of the proposed Multi-Attention Fusion Network (MAFNet) for RGB-T crowd counting. As shown in Fig. \ref{structure}, the MAFNet consists of an encoder part embedded with Multi-Attention Fusion (MAF) modules for cross-modal fusion and a Multi-modal Multi-scale Aggregation (MMA) regression head to make use of the cross-modal and contextual information for crowd density map prediction. We first introduce the encoder architecture of the MAFNet and the MAF module, which is the main component of the encoder part. Then, we describe the MMA regression head. Finally, the loss function is reported. 

\subsection{Encoder Architecture}
\label{encoder}
In the encoder part, the VGG19 \cite{simonyan2014very} networks which discard the fully-connected layers are used as the backbones to extract multi-scale features of the RGB and thermal modalities. The backbones are divided into 5 stages and the last layer of each stage is the max-pooling layer with a stride of 2 for down-sampling. To fully interact information of the two modalities, an intermediate fusion strategy is adopted by embedding the proposed MAF modules into the adjacent stages of the backbones. Considering that if the information interaction of large-scale features between the two modalities is carried out at the shallow stages, the fusion effect may be affected by the misalignment of the paired RGB-T data. In our implementation, the MAF modules are embedded between deep stages to fuse the features with small-scale but at the high semantic level, which alleviate the data misalignment problem through the translation invariance of the max-pooling layer \cite{liu2021cross} in the feature extraction stage. More specifically, the RGB and thermal images are fed to the two backbones respectively for modality-specific representation learning in stage 1 and 2. Starting from stage 3, the features are first further extracted through the backbones, and then fed to the MAF modules for information interaction and enhancement. Finally, three paired feature maps from the two modalities are obtained from the output of MAF blocks. Their scales are $1/8$, $1/16$, $1/32$ of the input images, and the corresponding number of channels are 256, 512, 512, respectively. 
% Formally:

% \begin{gather}
% \label{MAF_process}
%   \begin{cases}
%     F_r^{i-1},F_t^{i-1} = STG_i(F_r^{i-1},F_t^{i-1}), &\text{if $i=1,2$}\\
% F_r^i,F_t^i = MAF_i(STG_i(F_r^{i-1},F_t^{i-1})), &\text{if $i=3,4,5$}
%   \end{cases}
% \end{gather}
% where $STG_i$ means the $i$-th ($i=1,2,3,4,5$) stage of the five stages of backbones. $MAF_i$ is the MAF module embedded after stage $i$. $F_r^i-1$ and $F_t^i-1$ are the feature maps send into the $i$-th stage. 

\begin{figure*}[!t]
\centering
\includegraphics[width=0.9\textwidth]{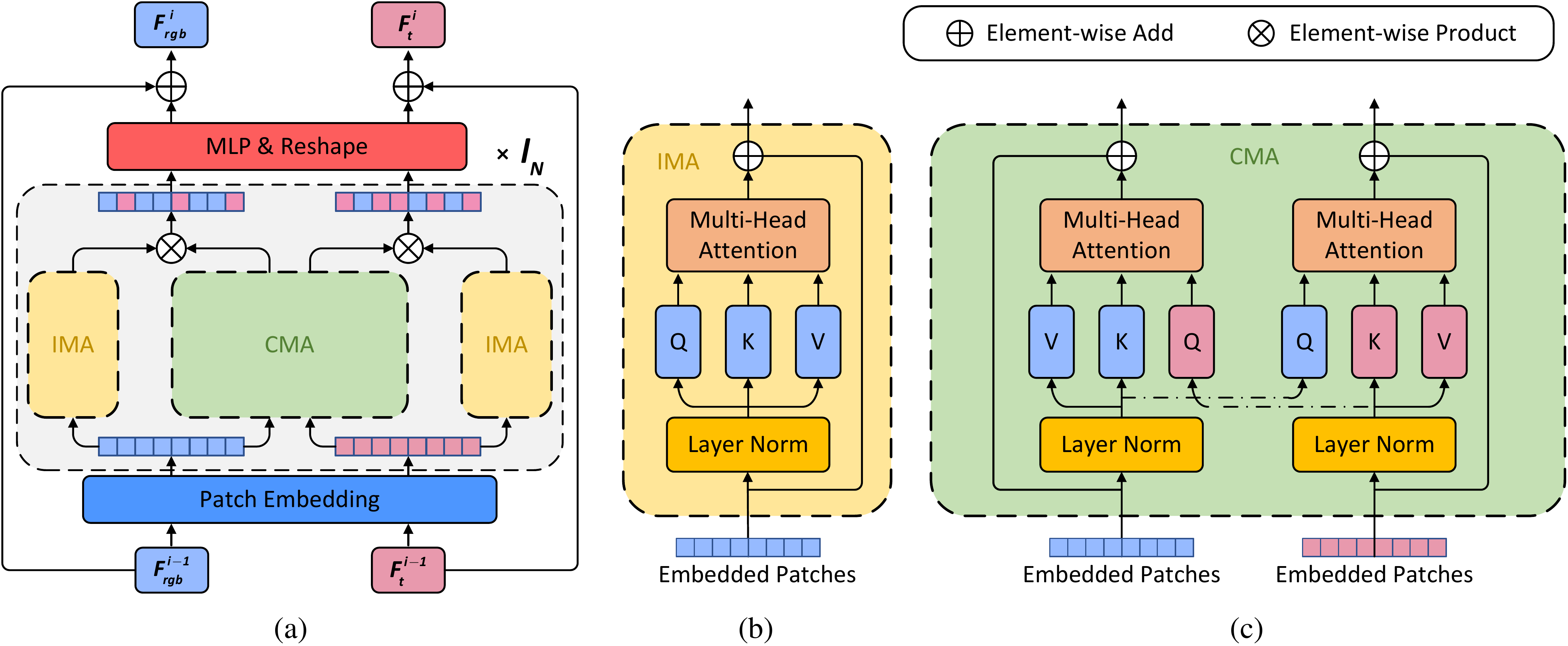}
\caption{\textbf{(a) MAF module}: stacking the MAF blocks which consist of Intra-Modality-Attention (IMA) module and Cross-Modality-Attention (CMA) module. \textbf{(b) IMA module}: further extracting intra-modal features. \textbf{(c) CMA module}: extracting cross-modal features.}
\label{MAF}
\end{figure*}

\subsection{Multi-Attention Fusion Module}
On account of the small perception field of convolution operations, the existing CNN-based multi-modal fusion modules have difficulty in adequately capturing long-range contextual information from the RGB and thermal modalities. Considering the powerful cross-modal global modeling capability of the attention mechanism, an attention-based Multi-Attention Fusion (MAF) module is proposed for better information interaction between the two modalities. The implementation of the MAF module represents in Fig. \ref{MAF}-(a).

\textbf{Patch Embedding}. To use the attention mechanism, the input 2D feature maps should first be converted into a sequence consisting of patches. Given a feature map $F_m \in \mathbb{R}^{C \times H \times W}$ of modal $m$ with the size of channel $C$, height $H$ and width $W$ and reshaped it into a sequence composing of $N$ patches of size $P$ with the dimension of $P^2\times C$, where $(N=HW/P^2)$. Then the patches are flattened and mapped to D-dimensional latent vectors via a trainable linear projection layer. The outputs of the projection layer are called patch embeddings, which can be formulated as:
\begin{equation}
\label{patch_embedding}
z_m^0 = [x_p^1E; x_p^2E; ...;x_p^NE],
\end{equation}
where $x_p^i$ means the $i$-th flattened patch ($i=1,2,...,N$) and $E$ $(E \in \mathbb{R}^{(P^2 \times C) \times D})$ denotes the linear projection matrix of the embedding process. In our implementation, the patch size of the first MAF module is set to 2 and the last two are set to 1. The embedding dimension $D$ is set to 768.

\textbf{MAF block}. To fully capture the long-range information within each modality and across modalities by the attention mechanism, the MAF block consisting of Intra-Modality-Attention (IMA) module and Cross-Modality-Attention (CMA) module is proposed. By stacking a suitable number of MAF blocks, the MAF module have a powerful cross-modal fusion capability at the global level. In this part, we first review the Multi-Head Attention mechanism and then introduce the implementation of the IMA and CMA modules. 

 \textbf{(a) Multi-Head Attention mechanism.} First, we introduce the implementation of Multi-Head Attention mechanism \cite{vaswani2017attention}. Given two patch embeddings $z_1$ and $z_2$ as inputs. Multi-Head Attention executes multiple independent attention heads in parallel and concatenates the outputs, which is then projected to obtain the final value. The input of one attention head $h$ is a triplet $Q_1^h$(query),  $K_2^h$(key) and $V_2^h$(value) calculated from the input embeddings:
\begin{align}
\label{MHA_QKV}
Q_1^h = z_1W^h_q, \notag \\
K_2^h = z_2W^h_k,        \\
V_2^h = z_2W^h_v, \notag
\end{align}
where $W^h_q$, $W^h_k$, $W^h_v \in \mathbb{R}^{D \times d}$ are the three learnable linear projection matrices and $d$ is the dimension of the channels. And one attention head is formulated as:
\begin{equation}
\label{HA}
Y^h = Attn(Q_1^h,K_2^h,V_2^h) = softmax(\frac{Q_1^h{K_2^h}^T}{\sqrt{d}})V_2^h,
\end{equation}
where $Y^h$ terms the output of one attetnion head. Then the Multi-Head Attention is expressed by the following formula:
\begin{equation}
\label{MHA}
z = Concat[Y^1, ..., Y^h]_{h=1}^{N_h}W^O,
\end{equation}
where $z$ is the output of the Multi-Head Attention processing. $N_h$ is the number of attention heads, $Concat[...]_{h=1}^{N_h}$ means the concatenation operation for all attention heads and $W^O \in \mathbb{R}^{hd \times N}$ is the linear projection matrices. In summary, the processing of the Multi-Head Attention is presented as follows:
\begin{equation}
\label{SMHA}
z = MHA_{h=1}^{N_h}(Q_1^h,K_2^h,V_2^h).
\end{equation}

\textbf{(b) Intra-Modality-Attention (IMA) module and Cross-Modality-Attention (CMA) module.} Based on the multi-head attention mechanism, the IMA and the CMA module are designed as shown in Fig. \ref{MAF}-(b, c), respectively. Specifically, in the $l$-th MAF block $(l=1,2,..,l_N)$, given the input patch embeddings $z_r^{l-1}$ and $z_t^{l-1}$ of different modalities. The input query, key and value of the IMA module are from the same modality for the intra-modal features enhancement. Formally:
\begin{align}
\label{IMA_r}
\{z_r^{l-1}\}^i &= MHA_{h=1}^{N_h}(\{Q_r^h\}^i,\{K_r^h\}^i,\{V_r^h\}^i) + z_r^{l-1}  \notag \\
                &= IMA_{r\to r}(z_r^{l-1}), 
\end{align}
\begin{align}
\label{IMA_t}
\{z_t^{l-1}\}^i &= MHA_{h=1}^{N_h}(\{Q_t^h\}^i,\{K_t^h\}^i,\{V_t^h\}^i) + z_t^{l-1} \notag \\
                &= IMA_{t\to t}(z_t^{l-1}),
\end{align}
where $\{...\}^i$ indicates the variable using in IMA module. $IMA_{r\to r}$ means the operation is performed within the RGB modality and $IMA_{t\to t}$ means the operation is performed within the thermal modality. Different from the IMA module, the input key and value of the CMA module are from the same modality but the query is from the other. The CMA module pays attention to the regions of interest across modalities for feature alignment and cross-modal information interaction by means of attention maps computed by the query and key of different modalities. Formally:
\begin{align}
\label{CMA_tr}
\{z_r^{l-1}\}^c &= MHA_{h=1}^{N_h}(\{Q_t^h\}^c,\{K_r^h\}^c,\{V_r^h\}^c) + z_r^{l-1}  \notag \\
                &= CMA_{t\to r}(z_r^{l-1}), 
\end{align}
\begin{align}
\label{CMA_rt}
\{z_t^{l-1}\}^c &= MHA_{h=1}^{N_h}(\{Q_r^h\}^c,\{K_t^h\}^c,\{V_t^h\}^c) + z_t^{l-1} \notag \\
                &= CMA_{r\to t}(z_t^{l-1}),
\end{align}
where $\{...\}^c$ indicates the variable using in CMA module. $CMA_{t \to r}$ means that the query is obtained from thermal modality, and key and value are obtained from RGB modality for co-attention. In contrast, $CMA_{r \to t}$ means that the query is from RGB modality, and key and value are from thermal modality. Then, the features obtained from different sub-modules are aggregated by element-wise product in different modalities. The aggregated process can be formulated as:
\begin{gather}
\label{agg}
z_r^l = \{z_r^{l-1}\}^i \odot \{z_r^{l-1}\}^c, \notag \\
z_t^l = \{z_t^{l-1}\}^i \odot \{z_t^{l-1}\}^c. 
\end{gather}

Finally, the features obtained from the last MAF block of a MAF module termed $z_r^{l_N}$ and $z_t^{l_N}$, where $l_N$ means the number of MAF blocks in one MAF module. In our implementation, the $l_N$ in each MAF module is taken as 2, i.e. stacking two MAF blocks.

\textbf{Recovering 2D structure and skip connection}. In order to obtain the 2D feature maps for the next stage of feature extraction, the final output features of the last MAF module should be mapped back to the dimension of the input embeddings by a linear projection layer and recovered the 2D structure. Besides, in consequence of the internal complexity of the MAF module may degrade the performance of the network, a skip connection structure is introduced to obtain the final output feature map of this stage by directly summing the input and output feature maps inspired by ResNet \cite{he2016deep}.

\subsection{Multi-Modal Multi-Scale Aggregation Regression Head} Considering that the output feature maps from the encoder may lose crucial information in the process of down-sampling and the feature maps with different scales may focus on different levels of semantic information from different modalities, a Multi-modal Multi-Scale Aggregation (MMA) regression head is proposed to aggregate the multi-modal and multi-scale feature information to reconstruct detail information and generate high-quality crowd density maps. Specifically, the paired feature maps of the two modalities with the same scale output from the the MAF modules are first concatenated by channel and then fed to the convolutions of $3\times 3$ to fully aggregate information of RGB and thermal modalities. Then the multi-scale features are up-sampled to the same scale and added together. Following, the added features are fed into four parallel branches, three of which contain dilated convolutions of $3\times 3$ with dilation rates of 1,2,3 for increasing the perceptive fields and one skip connection with  $1\times 1$ convolution. The output features of the top three branches are concatenated and added with the skip connection together to obtain the final aggregated feature maps. Finally, the aggregated feature maps are fed to a  $3\times 3$ convolution layer and a $1\times 1$ convolution layer to generate the predicted crowd density map. 

\subsection{Loss Function}
For training the proposed model, we adopt the standard Mean Squared Error (MSE) loss function, which uses Euclidean distance to calculate the gap between the ground truth and the predicted crowd density map. The loss function can be formulated as:
\begin{equation}
\label{loss}
L = \frac{1}{N}\sum\limits_{i = 1}^N {||{\hat{D}_i} - {D_i}||_2^2},
\end{equation}
where $N$ is the number of training samples, $i$ means the $i$-th training sample, $\hat{D}_i$ represents the predicted density map and $D_i$ represents the density map generated by ground truth.

\section{EXPERIMENTAL RESULTS AND ANALYSIS}
The experiments are implemented on two RGB-T crowd-counting datasets, RGBT-CC \cite{liu2021cross} and DroneRGBT \cite{zhang2021mmccn}. At the same time, the ablation studies are performed on the RGBT-CC dataset to demonstrate the effectiveness of the proposed network. Finally, the further analyses are discussed.

\begin{table*}[htbp]
\renewcommand\arraystretch{1.25}
    \normalsize
    \centering
    \caption{ABLATION STUDIES ON RGBT-CC DATASET.}
        \resizebox{0.7\textwidth}{!}{\begin{tabular}{c|cIcIc|c|c|c|c}
			\whline
			\multicolumn{2}{cI}{MAF} &\multirow{2}{*}{MMA}  &\multirow{2}{*}{GAME(0)} &\multirow{2}{*}{GAME(1)} &\multirow{2}{*}{GAME(2)} &\multirow{2}{*}{GAME(3)} &\multirow{2}{*}{RMSE} \\
			\cline{1-2}
			IMA	 &CMA  & & & & & & \\
			\whline
			\xmark  &\xmark &\xmark &14.55 &19.75	&24.55	&34.13	&28.37 \\
			\cline{1-8}
			\rmark  &\xmark &\xmark &13.13 &18.34	&23.11	&33.00	&24.67 \\
			\cline{1-8}
			\xmark  &\rmark &\xmark &12.50 &16.16	&20.55	&26.99	&22.30 \\
			\cline{1-8}
			\xmark  &\xmark &\rmark &13.67 &16.26	&19.67	&25.80	&24.90  \\
			\cline{1-8}
			\rmark  &\xmark &\rmark &12.46 &15.29 &19.11  &24.54  &23.91  \\	
			\cline{1-8}
			\xmark  &\rmark &\rmark &11.68 &14.81	&18.78  &24.44  &22.19  \\
			\cline{1-8}
			\rmark  &\rmark &\xmark  &11.85	&15.66	&20.47	&28.14	&22.11  \\
			\cline{1-8}
			\rmark  &\rmark &\rmark &\textbf{10.90} &\textbf{14.44}	   &\textbf{18.36} &\textbf{24.01}	   &\textbf{20.82}\\	
			\whline			
	    \end{tabular}}
    \label{ablation}
\end{table*}

\subsection{Datasets} 
\textbf{RGBT-CC} \cite{liu2021cross} is a large-scale RGB-T crowd counting dataset for surveillance view, which contains 2030 pairs of RGB-T images with the size of 640×480 consisting of 138,389 pedestrians. Among the 2030 images, 1013 images are captured under light conditions and 1017 images are captured under dark conditions.  

\textbf{DroneRGBT} \cite{peng2020rgb} is a RGB-T crowd counting dataset for drone view, which contains 3600 pairs of RGB-T images with the size of 640×512 consisting of 175,698 pedestrians. Among them, about 1600 images were captured under dusk conditions, 1300 images under light conditions, and 900 images under dark conditions.

\subsection{Experimental Setup}
\textbf{Implementation details.} The experiments are implemented on NVIDIA GeForce RTX 3090 GPU with 24GB memory. In the training phase, we adopt an optimizer of AdamW, the initial learning rate of $5 \times 10^{-5}$ with a linear warmup strategy, a batch size of 16 and a maximum number of iterations of 300. Normal data augmentation is applied to the input images, including rescale, random crop and flip. The input images are randomly cropped to the size of 256$\times$256. To convert the point annotations into the corresponding crowd density maps, we use Gaussian convolution with kernel size of 7 and variance $\sigma$ of 2 to extend the sparse points. In addition, our code is built on $C^3$-Framework \cite{gao2019c}.

\textbf{Evaluation Metrics.} We use Mean Absolute Error (MAE) and Root Mean Square Error (RMSE) as evaluation metrics in the two datasets. MAE and RMSE are defined as:
\begin{equation}
\label{MAE}
MAE = \frac{1}{N}\sum\limits_{i = 1}^N {|{\hat{P}_i} - {P_i}|},
\end{equation}
\begin{equation}
\label{RMSE}
RMSE = \sqrt {\frac{1}{N}\sum\limits_{i = 1}^N {{{({\hat{P}_i} - {P_i})}^2}} },
\end{equation}
where $\hat{P}_i$ and $P_i$ refer to the predicted count and ground truth of the $i$-th RGB-T image pairs, respectively. In addition, following \cite{liu2021cross}, Grid Average Mean Absolute Error (GAME) is utilized in the RGBT-CC dataset to evaluate the performance of different regions. Specifically, for a particular level $l$, the image is grided into $4^l$ non-overlapping regions to calculate the counting error of each region and sum up. GAME is defined as:
\begin{equation}
\label{GAME}
GAME(l) = \frac{1}{N}\sum\limits_{i = 1}^N {\sum\limits_{j = 1}^{{4^l}} {|\hat{P}_i^j - P_i^j|} },
\end{equation}
where $\hat{P}_i^j$ and $P_i^j$ refer to predicted count and ground truth of the $j$-th part of the $i$-th RGB-T image pairs, respectively. It is worth noting that GAME(0) corresponds to the counting error of the whole image, which is equivalent to the MAE. 

\subsection{Ablation Studies}
In order to verify the effectiveness of each component in the proposed Multi-Attention Fusion Network (MAFNet), extensive ablation studies are performed on the RGBT-CC dataset. A model with a dual VGG19 encoder and a regression head consisting of a $3\times 3$ convolution and a $1\times 1$ convolution is used as the baseline. We gradually add the proposed modules to the baseline to obtain variant models and compare the performance to the baseline to verify the effectiveness of each module. The evaluation results are shown in Table \ref{ablation}, and row 1 shows the result of the baseline.

\textbf{Effectiveness of Multi-Attention Fusion module.} 
The Multi-Attention Fusion (MAF) module is stacked by MAF blocks, which consists of two Intra-Modality-Attention (IMA) modules and a Cross-Modality-Attention (CMA) module. To demonstrate the effectiveness of the MAF module, we perform ablation studies both on the IMA and CMA modules. For results of the row 2 and row 3, the models with the IMA and CMA modules, respectively, both perform better than the baseline in all evaluation metrics, which verify the effectiveness of the IMA and the CMA module. Especially, the performance improvement of the model with the CMA module is much higher than that of the model with the IMA module. That's because the cross-modal information interaction relies mainly on the CMA module, which plays a more important role in the MAF module. Compared with the results in row 5 and row 6, the model with CMA also performs better than the model with IMA, further illustrating the importance of CMA in cross-modal information interaction. As shown in row 7, the model with both IMA and CMA modules exhibits satisfactory performance, which is close to the performance of the full structure of MAFNet (row 7 \emph{v.s.} row 8).

\begin{figure*}[!t]
\centering
\includegraphics[width=0.9\textwidth]{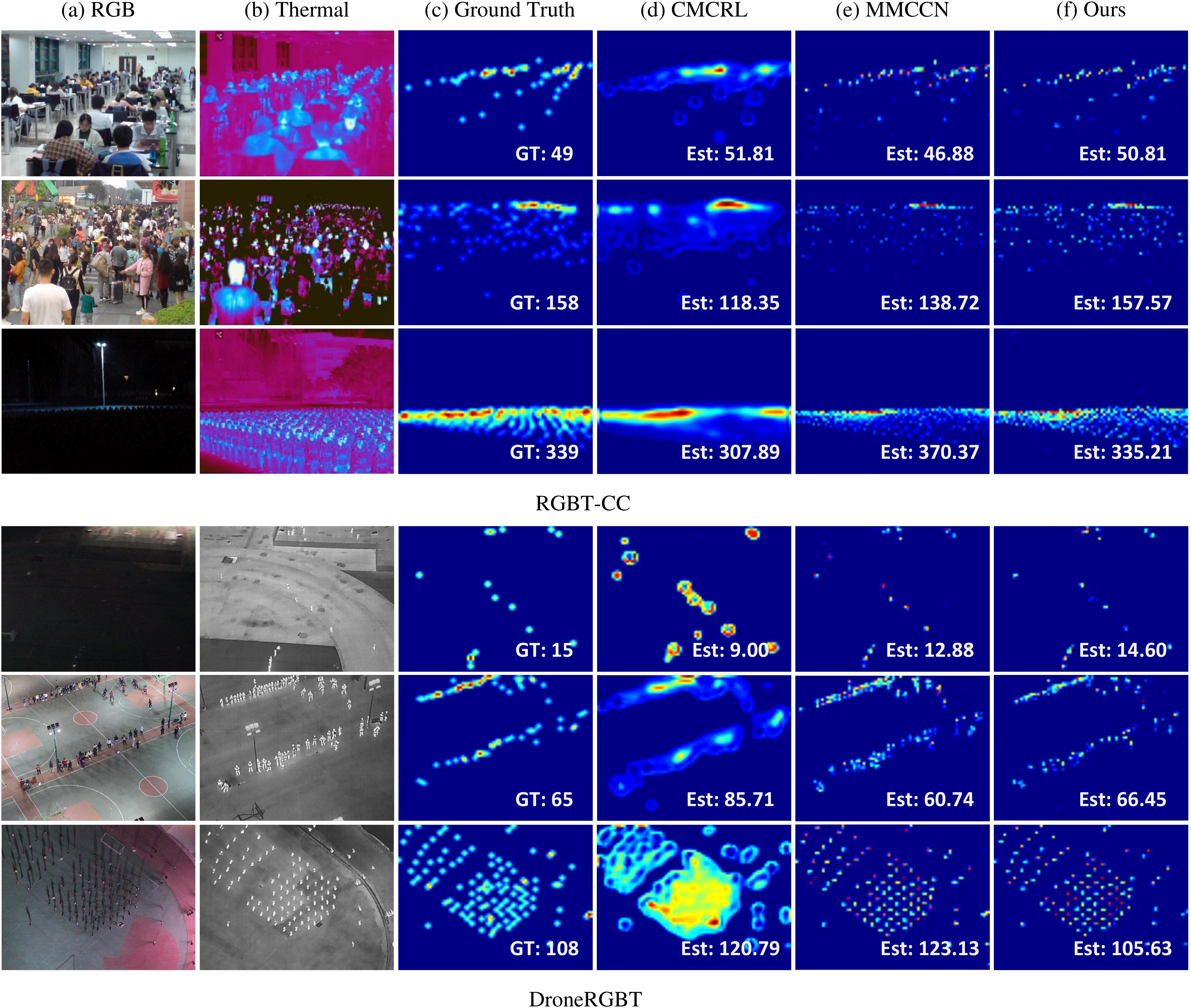}
\caption{Visualization of the crowd density maps of different methods on RGBT-CC and DroneRGBT datasets. The first three rows are the results of the RGBT-CC dataset and the last three rows are the results of the DroneRGBT dataset. (a) and (b) show the input RGB images and thermal images. (c) shows the ground-truths. (d) and (e) are the results of CMCRL \cite{liu2021cross} and MMCCN \cite{peng2020rgb}. (f) shows the results of our MAFNet. It could be observed that our predicted counts and density maps are more accurate than those of other state-of-the-art methods.}
\label{vis}
\end{figure*}

\textbf{Effectiveness of Multi-modal Multi-scale Aggregation regression head.} The Multi-modal Multi-Scale Aggregation (MMA) regression head is proposed to aggregate the multi-modal and multi-scale feature information to predict crowd density maps. Compared with row 4 and the baseline shown in row 1, the performance of the model is improved after adding the MMA module to the baseline model, which verifies the effectiveness of the MMA regression head. Further, the performance improvement of the MMA module combined with the IMA and CMA modules respectively are explored as shown in row 5 and row 6. The performance of the models with IMA and CMA modules respectively are both greatly improved on all evaluation metrics with the addition of MMA regression head, where GAME (0) are improved 5.10$\%$ / 6.56$\%$ (row 2 \emph{v.s.} row 5 and row 3 \emph{v.s.} row 6). From this, it could be observed that the model using the IMA module has a better improvement than the model with the CMA module after adding the MMA regression head. Considering that IMA module cannot implement cross-modal information interaction in the encoder stage, the multi-modal information is aggregated in the regression stage by introducing MMA to compensate for the limitation of IMA. Through the above extensive ablation studies, the effectiveness of the MMA regression head in aggregating multi-modal and multi-scale information in MAFNet is well demonstrated.

\begin{table*}[htbp]
\renewcommand\arraystretch{1.25}
    \normalsize
    \centering
    \caption{COMPARISON OF OUR APPROACH WITH OTHER PROPOSED BASELINE ON TWO DATASETS.}
    \begin{subtable}[t]{0.65\textwidth}
    \centering
    \caption{On RGBT-CC dataset}
        \setlength{\tabcolsep}{1.4mm}{\begin{tabular}{cIc|c|c|c|c}
			\whline
			\multirow{2}{*}{Method}  &\multicolumn{5}{c}{RGBT-CC} \\
			\cline{2-6}
			&\textbf{GAME(0)} & GAME(1) & GAME(2) & GAME(3) & \textbf{RMSE} \\
			\whline
			CSRNet \cite{li2018csrnet}      &20.40 &23.58 &28.03 &35.51 &35.26\\
			\hline
			BL \cite{ma2019bayesian}          &18.70 &22.55 &26.83 &34.62 &32.67\\
			\hline
			CMCRL \cite{liu2021cross}    &15.61	&19.95	&24.69	&32.89	&28.18\\
			\hline
			TAFNet \cite{tang2022tafnet}	&12.38	&16.98	&21.86	&30.19	&22.45\\
			\whline
		    MAFNet w/o MMA         &11.85	&15.66	&20.47	&28.14	&22.11\\
			\hline	
			\textbf{MAFNet (Ours)}	&\textbf{10.90} &\textbf{14.44}	   &\textbf{18.36} &\textbf{24.01}	   &\textbf{20.82}\\
			\whline			
	    \end{tabular}}
	    \label{sota_rgbtcc}
    \end{subtable}
    \begin{subtable}[t]{0.3\textwidth}
    \centering
    \caption{On DroneRGBT  dataset}
	\setlength{\tabcolsep}{1.4mm}{\begin{tabular}{cIc|c}
			\whline
			\multirow{2}{*}{Method}  &\multicolumn{2}{c}{DroneRGBT} \\
			\cline{2-3}
			&\textbf{MAE} & \textbf{RMSE} \\
			\whline
			CMCRL \cite{liu2021cross} &11.41 &17.54 \\
			\hline
			CSRNet \cite{li2018csrnet} &8.91	&13.80 \\
			\hline
			MMCCN \cite{peng2020rgb}	&7.27	&11.45\\
			\hline
			I-MMCCN \cite{zhang2021mmccn}	&6.91	&11.26	\\
			\whline
			MAFNet w/o MMA          &6.55   &10.64\\
			\hline
			\textbf{MAFNet (Ours)}	&\textbf{6.40}	& \textbf{10.16}	\\
			\whline			
	    \end{tabular}}
	    \label{sota_drone}
    \end{subtable}
    \label{sota}
\end{table*}

\begin{table*}[htbp]
\renewcommand\arraystretch{1.25}
    \normalsize
    \centering
    \caption{COMPARISON WITH OTHER METHODS IN DIFFERENT ILLUMINATION ON RGBT-CC DATASET.}
        \setlength{\tabcolsep}{1.4mm}{\begin{tabular}{cIcIc|c|c|c|c}
			\whline
			Illumination                &Method    &GAME(0)   &GAME(1)   &GAME(2)   &GAME(3)   &RMSE \\
			\whline
			\multirow{3}{*}{Brightness} &CMCRL \cite{liu2021cross}     &20.36      &23.57      &28.49      &36.29      &32.57\\
			\cline{2-7}
                                        &TAFNet \cite{tang2022tafnet}    &15.57    &20.65	   &26.67	   &36.17	   &24.25\\
			\cline{2-7}
			                            &\textbf{Ours}      &\textbf{11.31}	   &\textbf{14.61}	   &\textbf{18.92}	   &\textbf{25.31}	   &\textbf{21.81}\\
			\whline                            
			\multirow{3}{*}{Darkness} &CMCRL \cite{liu2021cross}     &15.44	   &19.23	   &23.79	   &30.28	   &29.11\\
			\cline{2-7}
                                        &TAFNet \cite{tang2022tafnet}    &14.20	   &19.20	   &24.00	   &31.63	   &27.50\\
			\cline{2-7}
			                            &\textbf{Ours}      &\textbf{10.48} 	   &\textbf{14.26}	   &\textbf{17.79}	   &\textbf{22.67}	   &\textbf{19.75}\\
			\whline			
	    \end{tabular}}
    \label{illumination}
\end{table*}

\subsection{Comparison with the SOTAs on RGBT-CC Dataset}
Table \ref{sota} lists the performance comparison between the MAFNet and four state-of-the-art methods on two datasets respectively and it could be found that the MAFNet shows strong performance, and outperforms the existing methods. As shown in Table \ref{sota}-(a), on the RGBT-CC dataset, MAFNet improved 11.95$\%$, 14.96$\%$, 16.01$\%$, 20.47$\%$, and 7.26$\%$ in the five metrics of GAME(0), GAME(1), GAME(2), GAME(3) and RMSE respectively, compared with the state-of-the-art method TAFNet \cite{tang2022tafnet}. And as shown in Table \ref{sota}-(b), compared with I-MMCCN \cite{zhang2021mmccn}, MAFNet achieves 7.38$\%$ and 9.77$\%$ improvement on MAE and RMSE. In addition, existing methods such as CMCRL \cite{liu2021cross} and TAFNet \cite{tang2022tafnet} do not utilize multi-scale information, and the performance of these models are completely dependent on the proposed multi-modal fusion modules. In fairness, we use the MAFNet without MMA regression head to compare with the state-of-the-art methods (row 4 \emph{v.s.} row 5 in Table \ref{sota}-(a) and (b)) and the experiments show that the MAFNet without MMA regression head still outperforms the state-of-the-art methods, which demonstrates the superiority of the designed MAF module over the existing multi-modal fusion modules. A visualization comparison of the partial state-of-the-art methods with MAFNet regarding the generated density maps for different density levels on the two datasets are shown in Fig. \ref{vis}. The visualization results show that MAFNet could predict more accurate counting results and perform better in dense crowd scenes than the state-of-the-art methods. 

To further illustrate the excellent robustness of MAFNet for illumination conditions, the RGBT-CC test set of 800 images is divided into 406 brightness and 394 darkness images to test the model's performance in different illumination conditions. As the results shown in Table \ref{illumination}, comparing with CMCRL \cite{liu2021cross} and TAFNet \cite{tang2022tafnet}, the MAFNet outperforms these methods under different illumination conditions. It is worth noting that the MAFNet exhibits similar performance under different illumination conditions, while the other methods are much different. This phenomenon precisely shows that MAFNet is more robust to illumination variations than other methods.

\begin{figure}[!t]
\centering
\includegraphics[width=0.45\textwidth]{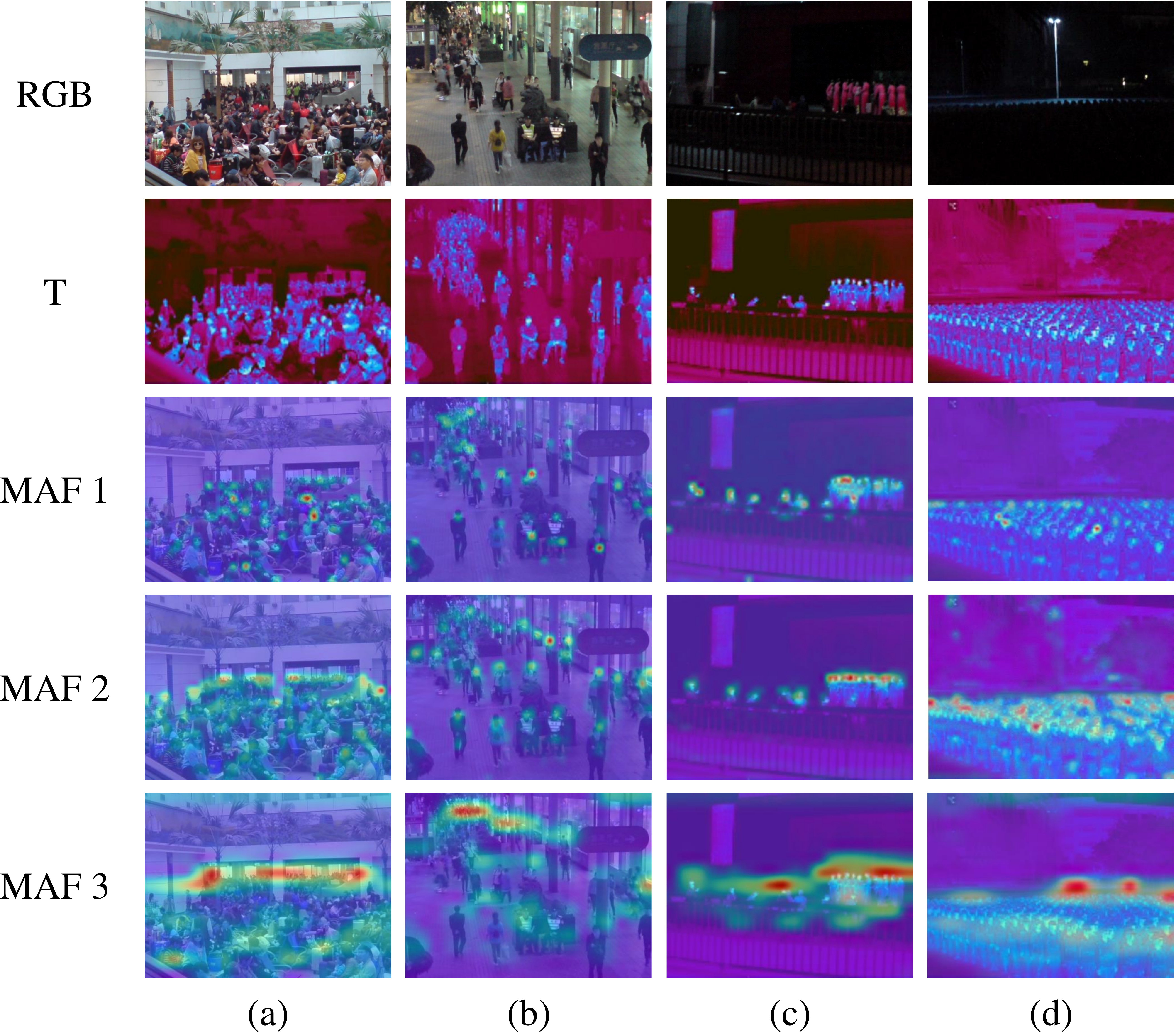}
\caption{Visualization of the attention maps from the three MAF modules. Red denotes higher attention values and blue denotes lower values.}
\label{attn_map}
\end{figure}

\subsection{Analysis and Discussion}

\begin{figure*}[!t]
\centering
\includegraphics[width=0.9\textwidth]{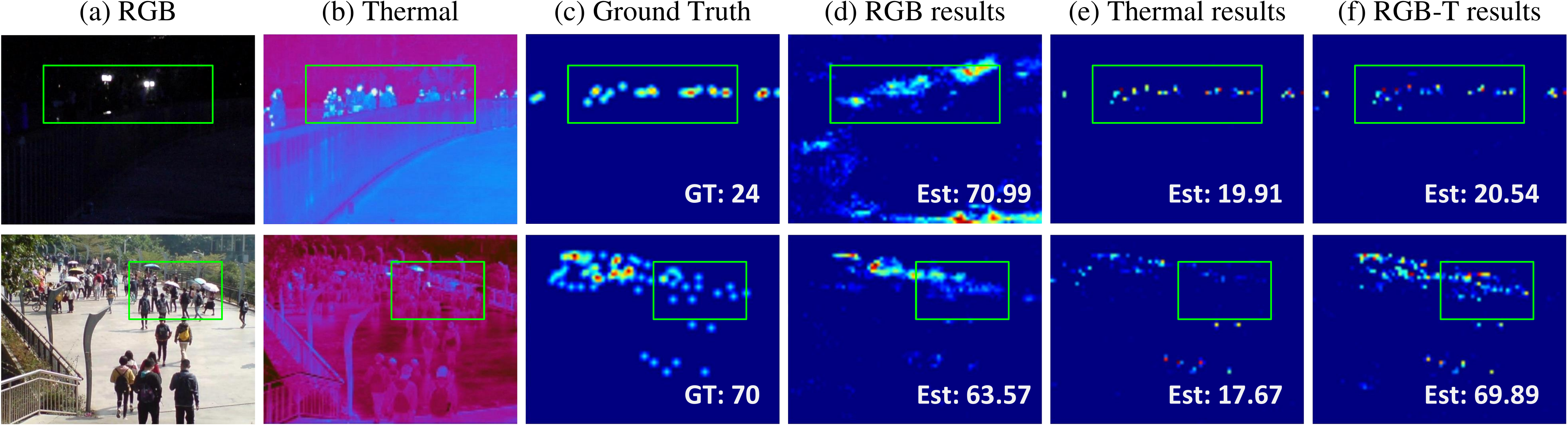}
\caption{Visualization results for generating crowd density maps using models trained with different modalities. (a) and (b) show the input RGB images and thermal images. (c) shows the ground-truths. (d) and (e) are the results of RGB-based MAFNet and thermal-based MAFNet. (f) shows  the results of MAFNet training on RGB-T data. And the ground-truths are shown in (g). The visualization results show that training the MAFNet with both RGB and thermal images leads to more accurate counting results and high-quality density maps.}
\label{abl_rgbt}
\end{figure*}

\textbf{Visualization analysis}.
To further illustrate the effectiveness of the proposed model, the attention maps of the three MAF modules are visualized in Fig. \ref{attn_map}. From this, it is clear that the network's attention is indeed focused on the target crowds, which illustrates the effectiveness of the attention mechanism used in the MAF module. Further, by comparing the attention maps obtained in different MAF modules, it could be observed that the MAF modules embedded at different locations focus on different levels of feature information. The shallow MAF module is better for head focus in sparse scenes because the inputs are large-scale feature maps, while it is difficult to fully focus on the entire crowd in more dense scenes. In contrast, the deep MAF module uses semantics-rich and small-scale feature maps that could model global information well for dense scene detection, but the performance of specific target recognition in sparse scenes decreases due to the low resolution. In particular, the second MAF module embedded in the middle stage has well-focused attention to crowds in different scales and density scenes. By observing the visualized results, it is a natural idea to introduce the MMA regression head to make full use of the multi-modal and multi-scale information for improving the robustness of the model to different density scenes.

\textbf{Effectiveness of RGB-T multi-modal data.} To verify the effectiveness of multi-modal data, RGB and thermal images are used to train the proposed model respectively and compared the performance with the RGB-T trained model. As shown in Table \ref{abl_rgbt}, using only RGB data to train the model works unsatisfactorily (row 1) because it is difficult to discriminate the people in RGB images under dark conditions, which provides insufficient information for crowd counting tasks. In comparison, the performance of the model trained with thermal images (row 2) is much better and closer to the performance of the model trained with RGB-T data (row 9), indicating that thermal images play a more important role in RGB-T crowd counting tasks, especially in dark environments. To further illustrate the necessity of multi-modal data, some extreme examples are listed in Fig. \ref{abl_rgbt}. In the first row, Fig. \ref{abl_rgbt}-(a) shows the RGB image with low-illumination condition, which leads to large error in the predicted crowd density map shown in Fig. \ref{abl_rgbt}-(d). In the last row, the thermal image shown in Fig. \ref{abl_rgbt}-(b) is poorly imaged with the unclear outline of the people in the boxed region leading to huge counting errors shown in Fig. \ref{abl_rgbt}-(e). Thanks to the high complementarity of RGB and thermal images, both of the modalities compensate for the information of the other to ensure the quality of the predicted crowd density maps, which are shown in Fig. \ref{abl_rgbt}-(f). 

\begin{table}[htbp]
\renewcommand\arraystretch{1.25}
    %\normalsize
    \centering
    \caption{EFFECTIVENESS OF RGB-T MULTI-MODAL DATA ON RGBT-CC DATASET.}
        \resizebox{0.45\textwidth}{!}{\begin{tabular}{cIc|c|c|c|c}
			\whline
			Input   &GAME(0)   &GAME(1)  &GAME(2)   &GAME(3)   &RMSE \\
			\whline
			RGB    &23.85      &30.18      &37.24      &45.40      &45.46\\
			\cline{1-6}
             T   &12.34    &15.50	   &19.25	   &24.45	   &22.84\\
			\cline{1-6}
			 RGB-T    	   &\textbf{10.90} &\textbf{14.44}	   &\textbf{18.36} &\textbf{24.01}	   &\textbf{20.82}\\
			\whline                            
	    \end{tabular}}
    \label{rgbt}
\end{table}

\textbf{Patch Size}. In the conventional vision transformer models, using large size patches such as 16×16 in ViT \cite{dosovitskiy2020image} will only result in coarse-grained features, which may cause large errors for intensive prediction tasks such as crowd counting. In contrast, using small size patches will make the sequence so long that it will cause a large amount of computation. However, as mentioned in Section \ref{encoder}, the MAF modules are only embedded in the high stages to process the small-scale features without causing much computation. Therefore, the patch size of the first MAF module is set to 2 and the last two are set to 1 to ensure both the acquisition of fine-grained features and low computation in our implementation.

\textbf{The necessity of Position Embedding.} Although position embedding is the default component of the attention-based transformer model \cite{vaswani2017attention}, it is worth discussing whether it is necessary in the MAF module. The comparison results of whether to add position embedding are shown in Table \ref{ablation} and $PE$ means the position embedding. It could be found that the two methods get comparable performance. Specifically, the network with position embedding performs poorly in GAME(0) (10.95 vs 10.90) but performs better in RMSE (19.85 vs 20.82). Considering that both convolution and attention mechanisms are used in the network, the skip connection of the MAF module somewhat introduces the inductive bias of convolution, which provides the location information for the attention calculation. Therefore, the addition of position embedding has little effect on the model's performance. 

\begin{table}[htbp]
\renewcommand\arraystretch{1.25}
    %\normalsize
    \centering
    \caption{EFFECTIVENESS OF POSITION EMBEDDING ON RGBT-CC DATASET.}
        \resizebox{0.45\textwidth}{!}{\begin{tabular}{cIc|c|c|c|c}
			\whline
			Method   &GAME(0)   &GAME(1)  &GAME(2)   &GAME(3)   &RMSE \\
			\whline
			w.  PE    &10.95      &14.41      &18.62      &24.48      &\textbf{19.85}\\
			\cline{1-6}
            w/o PE   &\textbf{10.90} &\textbf{14.44}	   &\textbf{18.36} &\textbf{24.01}	   &20.82\\
			\whline                            
	    \end{tabular}}
    \label{pe}
\end{table}

\textbf{Comparison of different numbers of MAF modules.}
The number of MAF modules is an important hyperparameter. Assuming that the MAF modules are also embedded in the earlier stages, on the one hand the computation will increase, and on the other hand the additional counting errors will be introduced due to the paired image unalignment, which has mentioned in Section \ref{encoder}. To find the optimal number of MAF modules, the MAF modules are embedded after each stage starting from the last stage of the backbone in turn. As shown in Table \ref{num}, the performance of the network is gradually getting better by increasing the number of MAF modules and the best result is obtained when the number of MAF modules is 3. When increasing to 4 modules, the model's performance starts to degrade. Eventually, the number of the MAF modules is set to 3 in the network.

\begin{table}[htbp]
\renewcommand\arraystretch{1.25}
    %\small
    \centering
    \caption{EFFECTIVENESS OF THE NUMBERS AND DEPTH OF THE MAF MODULES.}
        \resizebox{0.48\textwidth}{!}{\begin{tabular}{cIcIc|c|c|c|c}
			\whline
			Nums   &Depth   &GAME(0)   &GAME(1)   &GAME(2)   &GAME(3)  &RMSE \\
			\whline
			1      &2,2,2                          &11.88	&15.33	&19.14	&26.25	&23.04\\
			\whline
            2      &2,2,2                          &11.55	&15.12	&19.24	&26.35	&21.14\\
			\whline
			\multirow{4}{*}{3}      &1,1,1         &12.42 	&15.43	&18.97	&24.08	&22.09\\
			\cline{2-7}
			                        &1,2,4         &12.08 	&15.45	&19.33	&26.56	&22.87\\
			\cline{2-7}
			                        &4,2,1         &11.67 	&14.54	&\textbf{18.05}	&\textbf{23.33}	&21.08\\
			\cline{2-7}
			                        &2,2,2         &\textbf{10.90} &\textbf{14.44}	   &18.36 
			            &24.01	   &\textbf{20.82}\\
			\whline
            4       &2,2,2    &12.22  	&15.09	&18.52	&23.87	&21.67\\
			\whline			
	    \end{tabular}}
    \label{num}
\end{table}

\textbf{Comparison of different depths of MAF module.}
The depth of MAF module means the number of stacked MAF blocks in MAF module, which is also an important hyperparameter. We test different configurations of the network that contains different depth of MAF modules and show the results in Table \ref{num}. The 3 numbers in the second column of Table \ref{num} correspond to the depth of each of the 3 MAF modules, which indicates the depth of the MAF module greatly affects the performance of the model. Considering that the performance of MAE(GAME(0)) and RMSE is more important than other evaluation metrics, we set the depth of all three MAF modules to 2 as our final implementation to get the best result.

\section{CONCLUSION}
In this paper, we propose an effective two-stream framework for RGB-T crowd counting termed Multi-Attention Fusion Network (MAFNet). By embedding the attention-based Multi-Attention Fusion (MAF) modules between different stages of the backbone, the perception field of network in the cross-modal fusion stage is expanded to fully capture the long-range contextual information across modalities. The proposed Multi-modal Multi-scale Aggregation (MMA) regression head integrates multi-scale information across modalities and further improves the network's performance. Extensive experiments show the effectiveness of the proposed method and the network achieves state-of-the-art performance on RGBT-CC and DroneRGBT datasets. The ablation studies demonstrate the effectiveness of the individual modules.

%\section*{ACKNOWLEDGMENT}

\bibliographystyle{IEEEtran}
\bibliography{IEEEabrv,reference}

\begin{IEEEbiography}[{\includegraphics[width=1in,height=1.25in,clip,keepaspectratio]{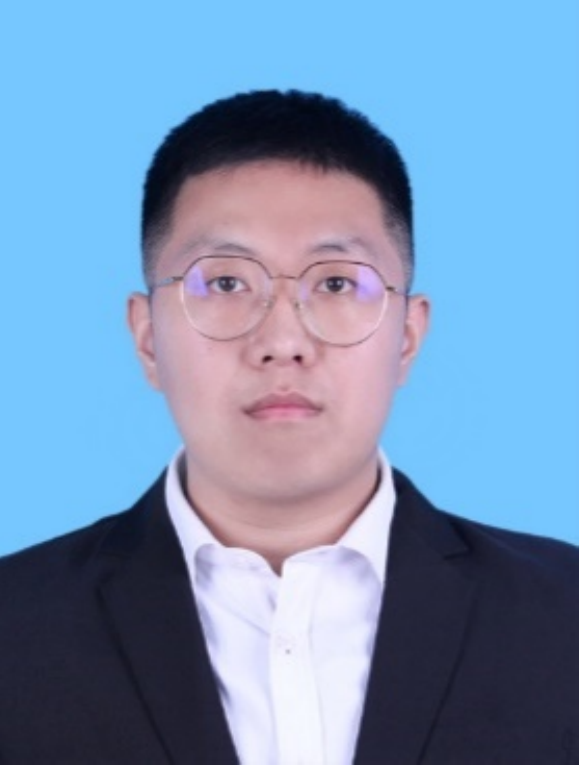}}]{Pengyu Chen} received the B.E. degree in Automation from Northwestern Polytechnical University, Xi'an 710072, Shaanxi, P. R. China, in 2022. He is currently pursuing the M.S. degree in computer science and technology with the School of Computer Science, and the school of Artificial Intelligence, Optics and Electronics (iOPEN), Northwestern Polytechnical University. His research interests include computer vision and pattern recognition.
\end{IEEEbiography}
\begin{IEEEbiography}[{\includegraphics[width=1in,height=1.25in,clip,keepaspectratio]{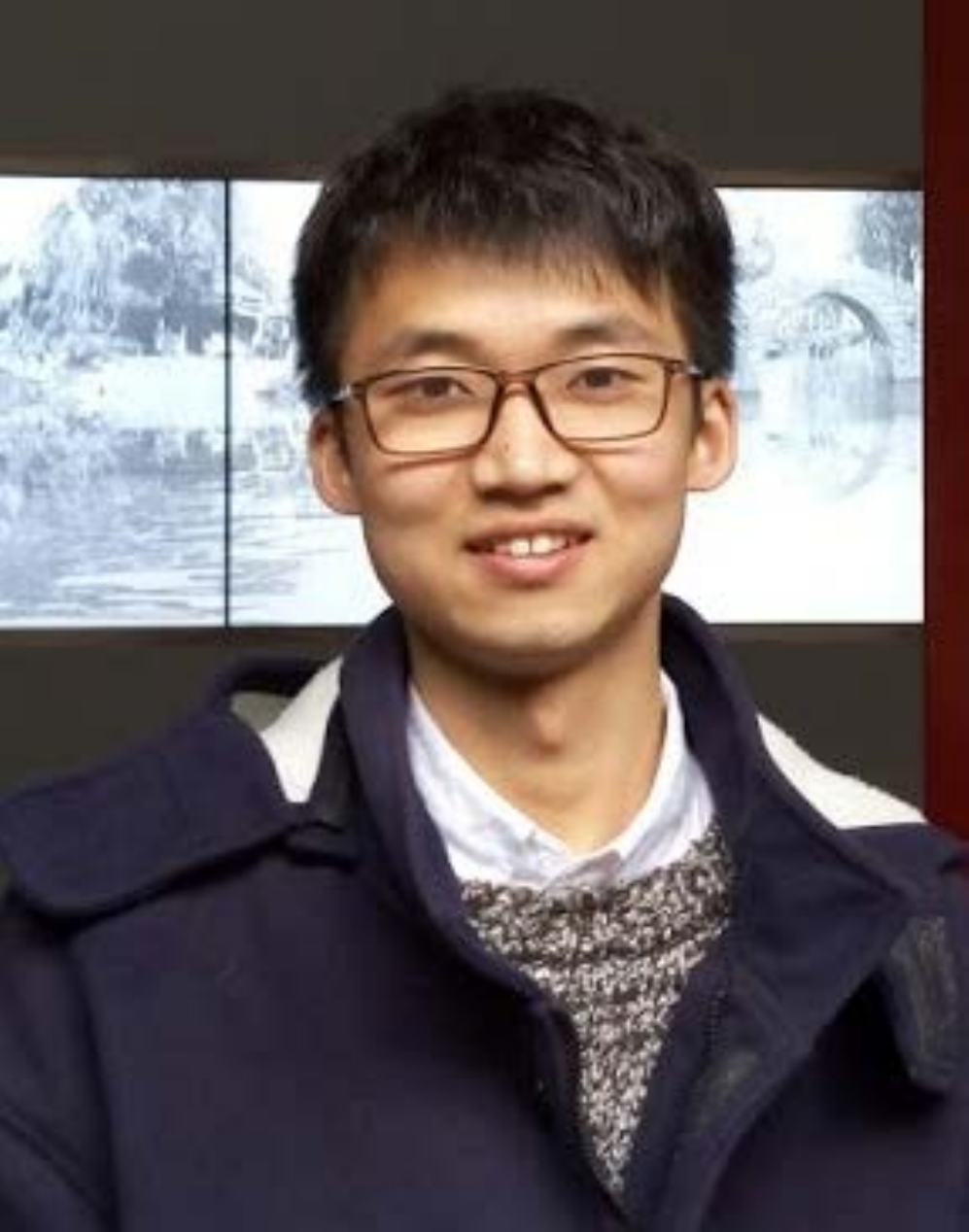}}]{Junyu Gao} received the B.E. degree and the Ph.D. degree in computer science and technology from the Northwestern Polytechnical University, Xi'an 710072, Shaanxi, P. R. China, in 2015 and 2021 respectively. He is currently a researcher with the School of Artificial Intelligence, Optics and Electronics (iOPEN), Northwestern Polytechnical University, Xi’an, China. His research interests include computer vision and pattern recognition.
\end{IEEEbiography}
\begin{IEEEbiography}[{\includegraphics[width=1in,height=1.25in,clip,keepaspectratio]{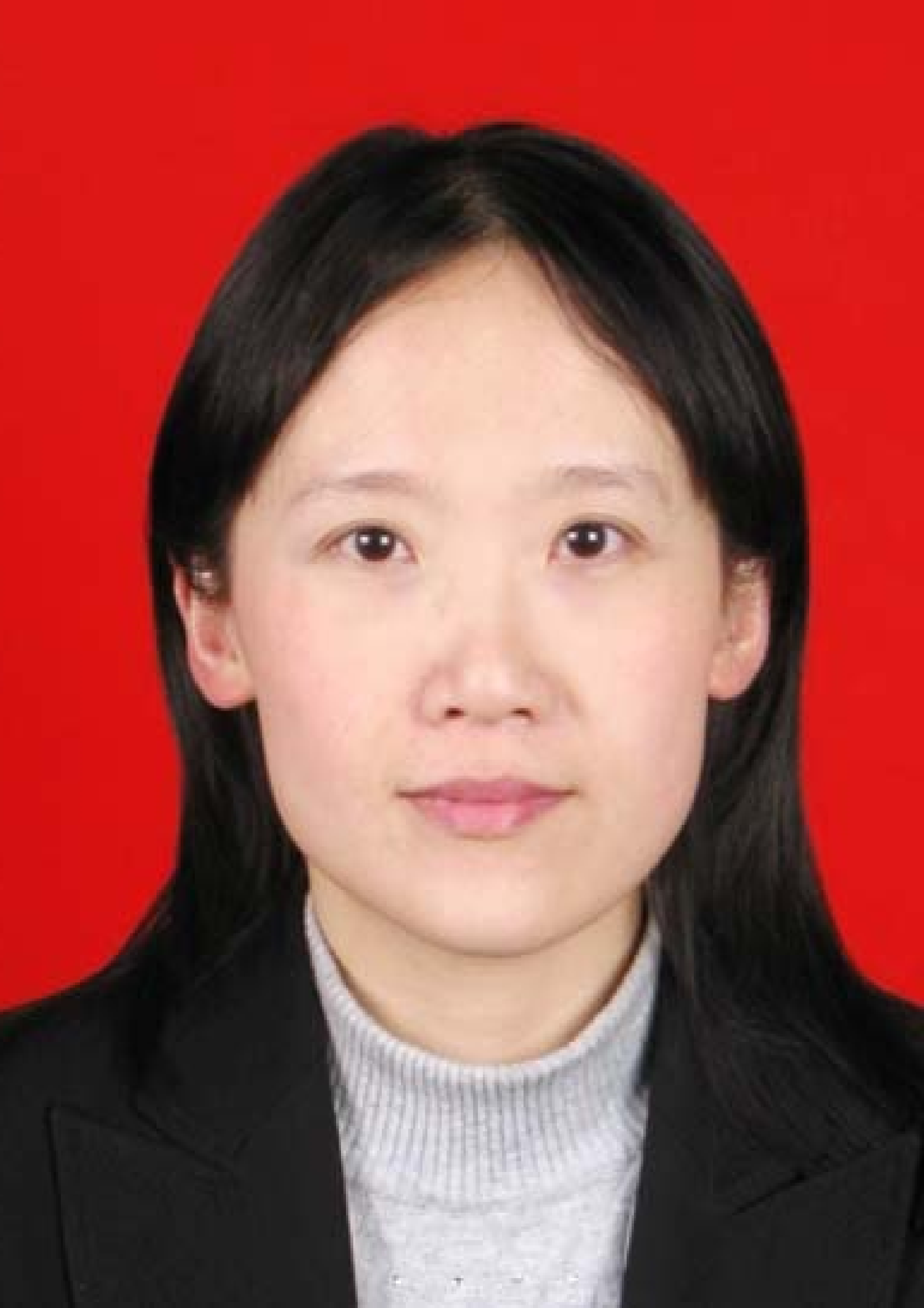}}]{Yuan Yuan}(M'05-SM'09) is currently a Full Professor with the School of Computer Science and the School of Artificial Intelligence, Optics and Electronics (iOPEN), Northwestern Polytechnical University, Xi'an, China. She has authored or coauthored over 150 papers, including about 100 in reputable journals, such as the
IEEE TRANSACTIONS ON PATTERN ANALYS IS AND MACHINE INTELLIGENCE, as well as the conference papers in IEEE Conference on Computer Vision and Pattern Recognition (CVPR), British Machine Vision Conference (BMVC), International Conference on Image Processing (ICIP), and International Conference on Acoustics, Speech and Signal Processing (ICASSP). Her current research interests include visual information processing and image/video content analysis.
\end{IEEEbiography}

\begin{IEEEbiography}[{\includegraphics[width=1in,height=1.25in,clip,keepaspectratio]{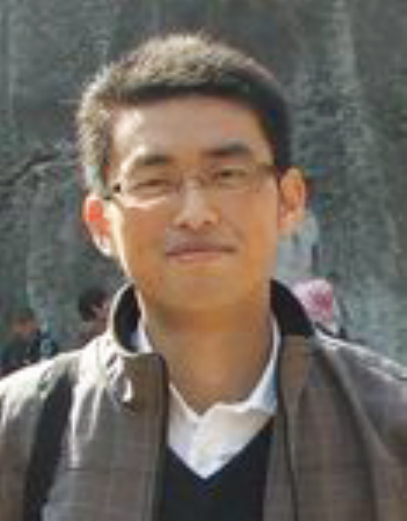}}]{Qi Wang} received the B.E. degree in automation and the Ph.D. degree in pattern recognition and intelligent systems from the University of Science and Technology of China, Hefei, China, in 2005 and 2010, respectively. He is currently a Professor with the School of Artificial Intelligence, Optics and Electronics (iOPEN), Northwestern Polytechnical University, Xi’an, China. His research interests include computer vision, pattern recognition, and remote sensing.
\end{IEEEbiography}

% \section{Additional Contents}
% The description of the PFA and MDC is short. Maybe the followling Table \ref{abl_decoder} should be added. Temporarily recorded here later to consider how to add it.

% \begin{table}[htbp]
%     \normalsize
%     \centering
%     \caption{Ablation study of different network compositions in the proposed method on RGBT-CC dataset.}
%         \setlength{\tabcolsep}{1.4mm}{\begin{tabular}{cIc|c|c|c|c}
% 			\whline
% 			Method    &GAME(0)   &GAME(1)   &GAME(2)   &GAME(3)   &RMSE \\
% 			\whline
% 			MAF Encoder     &11.63	&15.36	&19.73	&26.40	&21.68 \\
% 			\cline{1-6}
%             MAF Encoder+FPN+MSD  &12.37 &15.16 &18.48 &23.70 &23.05 \\
% 			\cline{1-6}
%             MAF Encoder+PFA+MSD' &12.27 &15.41 &19.29 &24.93 &24.42	\\
% 			\cline{1-6}
%             MAF Encoder+PFA+MSD    &10.75	&14.27	&18.08	&23.70	&20.68	\\
% 			\cline{1-6}
% 			\whline
% 	    \end{tabular}}
%     \label{abl_decoder}
% \end{table}

\end{document}